\documentclass[letterpaper]{article} 
\usepackage{aaai25}  
\usepackage{times}  
\usepackage{helvet}  
\usepackage{courier}  
\usepackage[hyphens]{url}  
\usepackage{graphicx} 
\usepackage{natbib}  
\usepackage{caption} 
\usepackage{algorithm}
\usepackage{algorithmic}

\usepackage{newfloat}
\usepackage{listings}
\floatstyle{ruled}
\newfloat{listing}{tb}{lst}{}
\floatname{listing}{Listing}
\usepackage[colorlinks]{hyperref}
\hypersetup{
    linkcolor=black,     
    citecolor={rgb,255:red,0;green,0;blue,90},   
    filecolor=magenta,  
    urlcolor=pink     
}

\usepackage{booktabs}
\usepackage{xspace}
\usepackage{stix,bbding,pifont,utfsym,fontawesome}
\usepackage{xcolor,colortbl}
\definecolor{citecolor}{HTML}{0071BC}
\definecolor{linkcolor}{HTML}{ED1C24}
\definecolor{ForestGreen}{HTML}{529578}
\definecolor{Bittersweet}{HTML}{c95863}

\usepackage{xcolor,colortbl}
\definecolor{citecolor}{HTML}{0071BC}
\definecolor{linkcolor}{HTML}{ED1C24}
\definecolor{ForestGreen}{HTML}{529578}
\definecolor{Bittersweet}{HTML}{c95863}

\newcommand{\system}{Prova\xspace}
\newcommand{\MissedData}{-\xspace}

\def\eg{\emph{e.g.}\xspace} 

\def\ie{\emph{i.e.}\xspace}

\usepackage{amsmath}
\usepackage[capitalize]{cleveref}
\crefname{section}{Sec.}{Secs.}
\Crefname{section}{Section}{Sections}
\Crefname{table}{Table}{Tables}
\crefname{table}{Tab.}{Tabs.}
\crefname{section}{\S}{\S\S}
\crefname{subsection}{\S}{\S\S}
\crefformat{table}{Table~#2#1#3}
\crefformat{figure}{Figure~#2#1#3}
\crefformat{equation}{Equation~(#2#1#3)}

\newcommand{\tablestyle}[2]{\setlength{\tabcolsep}{#1}\renewcommand{\arraystretch}{#2}\centering}
\newlength\savewidth
\usepackage{makecell}

\title{Comprehensive Multi-Modal Prototypes are Simple and Effective Classifiers\\
for Vast-Vocabulary Object Detection}
\author{
    Yitong Chen\textsuperscript{\rm1,\rm2}\equalcontrib,
    Wenhao Yao\textsuperscript{\rm 1}\equalcontrib,
    Lingchen Meng\textsuperscript{\rm 1}\equalcontrib,
    Sihong Wu\textsuperscript{\rm 1},
    Zuxuan Wu\textsuperscript{\rm1,\rm2}\thanks{Corresponding author.},
    Yu-Gang Jiang\textsuperscript{\rm 1}
}
\affiliations{
    \textsuperscript{\rm 1}Shanghai Key Lab of Intell. Info. Processing, School of CS, Fudan University, \textsuperscript{\rm 2}Shanghai Innovation Institute\\
    \{chenyt24, whyao23, shwu22\}@m.fudan.edu.cn, \{lcmeng20, zxwu, ygj\}@fudan.edu.cn
}

\usepackage{bibentry}

\begin{document}

\maketitle

\begin{abstract}
Enabling models to recognize vast open-world categories has been a longstanding pursuit in object detection. By leveraging the generalization capabilities of vision-language models, current open-world detectors can recognize a broader range of vocabularies, despite being trained on limited categories. However, when the scale of the category vocabularies during training expands to a real-world level, previous classifiers aligned with coarse class names
significantly reduce the recognition performance of these detectors. In this paper, we introduce \textbf{\system}, a multi-modal \textbf{pro}totype classifier for \textbf{va}st-vocabulary object detection. \system extracts comprehensive multi-modal prototypes as initialization of alignment classifiers to tackle the vast-vocabulary object recognition failure problem. On V3Det, this simple method greatly enhances the performance among one-stage, two-stage, and DETR-based detectors with only additional projection layers in both supervised and open-vocabulary settings. In particular, \system improves Faster R-CNN, FCOS, and DINO by 3.3, 6.2, and 2.9 AP respectively in the supervised setting of V3Det. For the open-vocabulary setting, \system achieves a new state-of-the-art performance with 32.8 base AP and 11.0 novel AP, which is of 2.6 and 4.3 gain over the previous methods.
\end{abstract}


\begin{links}
    \link{Code}{https://github.com/Row11n/Prova}
\end{links}

\section{Introduction}
The ability of humans to recognize a vast number of object categories in real-world scenarios has always been the goal of open-world object detection. Although fully supervised closed-set object detection methods have made significant progresses~\cite{fastrcnn, FCOS, normlinearlayer, deformable_detr, dino}, they still struggle to handle vast categories in the wild due to the limited number of object categories in commonly used detection datasets, \eg, COCO~\cite{coco} with 80 categories and Object365~\cite{objects365} with 365 categories. To address this issue, remarkable advancements have been achieved in an open-vocabulary manner which have acceptable zero-shot ability to unseen categories during training and also comparable closed-set capability~\cite{detic, regionCLIP}. Most of the recent open-vocabulary detectors leverage the generalization capabilities of Vision-Language Models (VLMs) like CLIP~\cite{clip}, by converting the closed-set prediction head into vision-language alignment between objects and class names.

\begin{figure}[!t]
\centering
\includegraphics[width=1.0\columnwidth]{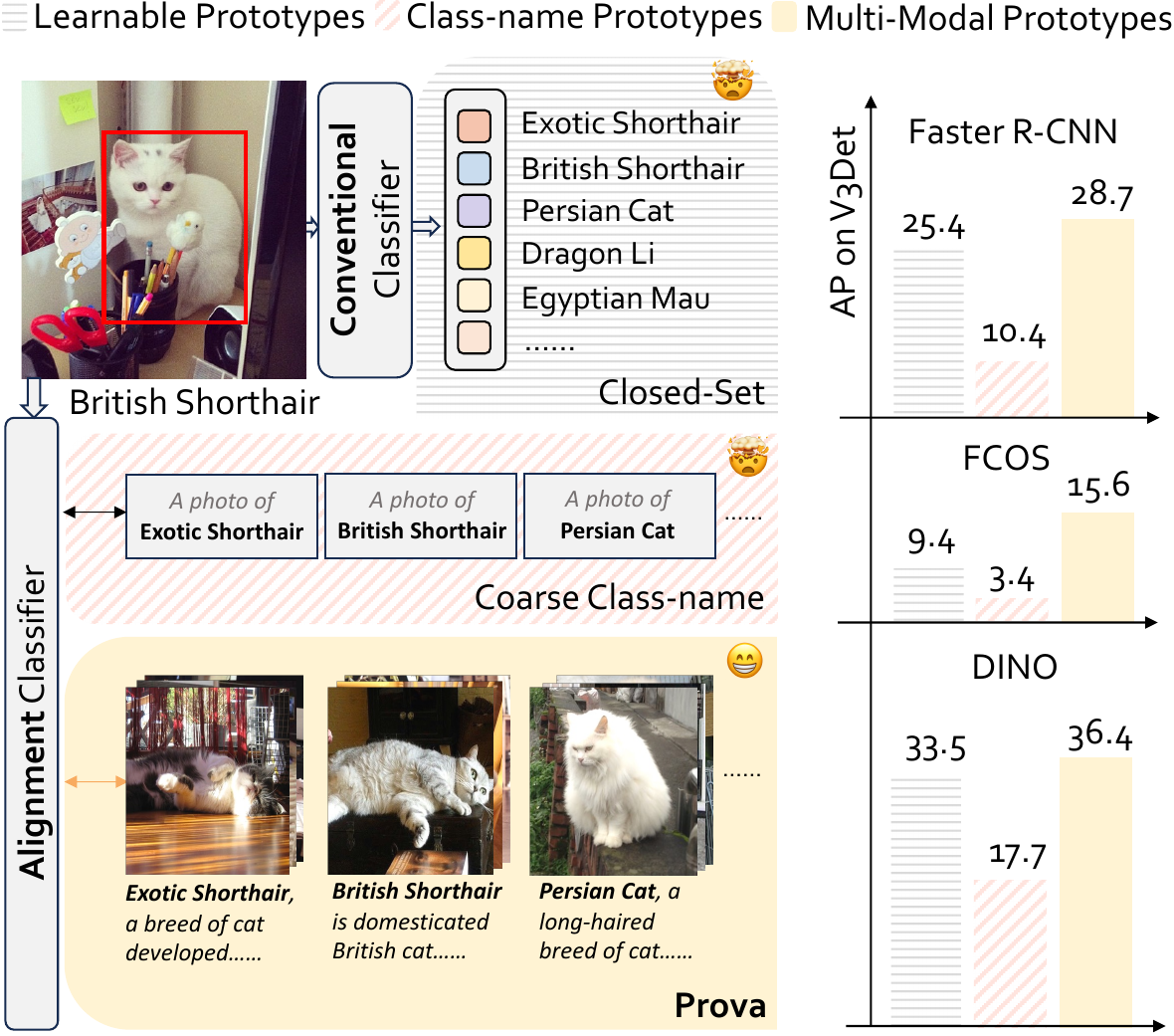} 
\caption{Left: When the category vocabularies scale up, previous conventional closed-set classifiers and class name alignment classifiers struggle to distinguish complex classes such as \textit{Exotic Shorthair} and \textit{British Shorthair}. Our key idea is to extract more detailed multi-modal prototypes to replace coarse class names with templates in the alignment classifier. Right: AP on V3Det validation set in a  supervised manner. The class name alignment classifier performs even worse than the conventional classifier, and multi-modal prototype classifier, \ie \system, achieves the best results.}
\label{fig:teaser}
\end{figure}

\begin{figure*}[!t]
\centering
\includegraphics[width=1\textwidth]{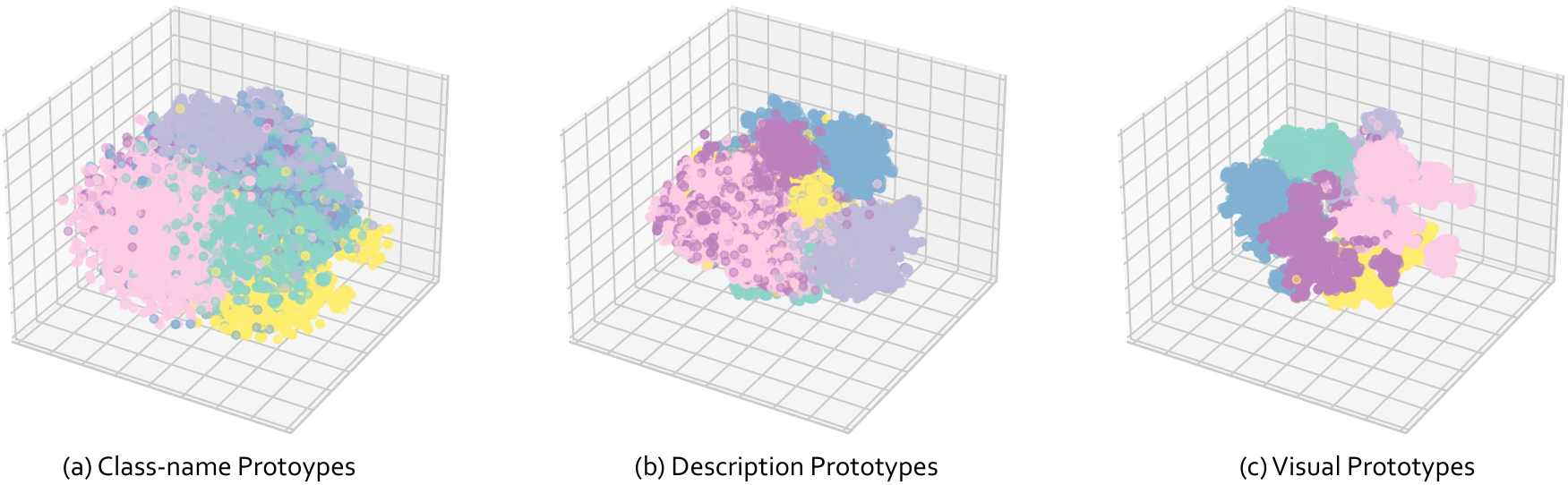} 
\caption{ Feature space visualization of (a) Class names with fixed templates; (b) Detailed class descriptions; (c) Reference images of each class. We use CLIP-ViT-Large~\cite{clip} as the text encoder to extract class name prototypes, LongCLIP-Large~\cite{longclip} to extract description prototypes, CLIP-ViT-Large visual encoder to extract visual prototypes and then utilize t-SNE~\cite{tsne} after $k$-means. Compared to class name prototypes, description and visual prototypes have a cleaner feature space, which is preferable for alignment classifier.}
\label{fig:visual}
\end{figure*}

However, as shown in~\cref{fig:teaser}, when the number of categories scales up in the detection dataset, \eg V3Det~\cite{v3det} with 13204 categories, the class name alignment classifier, which is well done in open-vocabulary fields, fails and performs even worse than the conventional closed-set classifier. 
Previous vision-language alignment methods assume a mapping relationship from detected regions to class vocabularies~\cite{vild,regionCLIP}, which holds when the vocabulary size is small, but becomes problematic as the vocabulary size expands. Specifically, 
coarse class names with fixed templates as the textual guidance encounter issues like semantic confusion (\eg, \textit{Exotic Shorthair} and \textit{British Shorthair} shown in~\cref{fig:teaser}) and ambiguity (\eg, \textit{Crane} in V3Det categories, which could be a machine or an animal belonging to Gruiformes), leading to difficulties in maintaining the mapping relation.

Furthermore, it is challenging even for a human to distinguish whether the cat detected in~\cref{fig:teaser} is an \textit{Exotic Shorthair} or a \textit{British Shorthair}. However, with detailed descriptions and reference images of both breeds, we can easily identify the correct result. Based on these observations, we propose \textbf{\system}, a multi-modal \textbf{pro}totypes classifier for \textbf{va}st-vocabulary object detection. Specifically, we utilize the detailed descriptions from the V3Det Challenge 2024~\cite{v3detcha} to generate description embedding via a text encoder. We refer to these description embeddings as textual prototypes to replace the original class name embedding in previous alignment classifiers. Additionally, we collect images from the training dataset and exemplars in V3Det to generate visual prototypes via a visual encoder. Finally, we utilize the textual and visual prototypes for classification through dot product alignment with detected region features.

As a proof-of-concept, shown in~\cref{fig:visual}, we visualize the feature space of different prototypes. It illustrates that more detailed text descriptions enable the identification of more accurate semantic points within the feature space of the language modality, thereby enhancing the differentiation between categories. In addition, the feature space of the visual modality demonstrates greater distinction and reduced ambiguity compared to the language modality. By integrating both modalities, the classifier can better align with the detected objects, leading to improved object recognition results. As quantitative proof, we conduct experiments in supervised and open-vocabulary settings on V3Det. The results show that our \system is compatible with various detection frameworks, significantly improving object recognition accuracy in both settings with only two additional projection layers and four additional matrix multiplications. 
Furthermore, our best model with a \textbf{base-size} backbone, achieves a new state-of-the-art performance in both settings, particularly being 5$\times$ lighter than the previous SoTA models with a \textbf{huge-size} backbone in the supervised setting. Our contributions can be summarized as:
\begin{itemize}
\item We point out that previous open-world classifiers using class names aligned with visual features are struggling when categories increase, and more detailed descriptions and reference images can alleviate this problem.
\item We propose a simple yet novel approach, \system, to extract multi-modal prototypes and leverage them to perform vast-vocabulary recognition, which is applicable to various detection frameworks with minimal modification.
\item Our method demonstrates efficient and powerful results on vast-vocabulary detection datasets, \eg, V3Det, by only two additional projection layers and four additional matrix multiplications.
\end{itemize}

\section{Related Work}

\subsection{Supervised Object Detection}
Supervised object detection models, in a closed-set setting where the category vocabularies remain constant during both training and inference, can be divided into three frameworks: two-stage, one-stage, and DETR-based. Two-stage detectors~\cite{fastrcnn, cascade, center, retina} first utilize a Region Proposal Network (RPN) to predict class-agnostic proposals that are pooled to region-of-interest features, and then recognize them by a classifier. One-stage detectors~\cite{FCOS, yolo, ssd} directly regress and classify the anchor boxes or perform dense searches for visual cues. DETR-based detectors~\cite{detr, deformable_detr, dino, dethub} treat object detection as a set prediction problem by utilizing an encoder-decoder transformer architecture.

\begin{figure*}[!ht]
\centering
\includegraphics[width=1\textwidth]{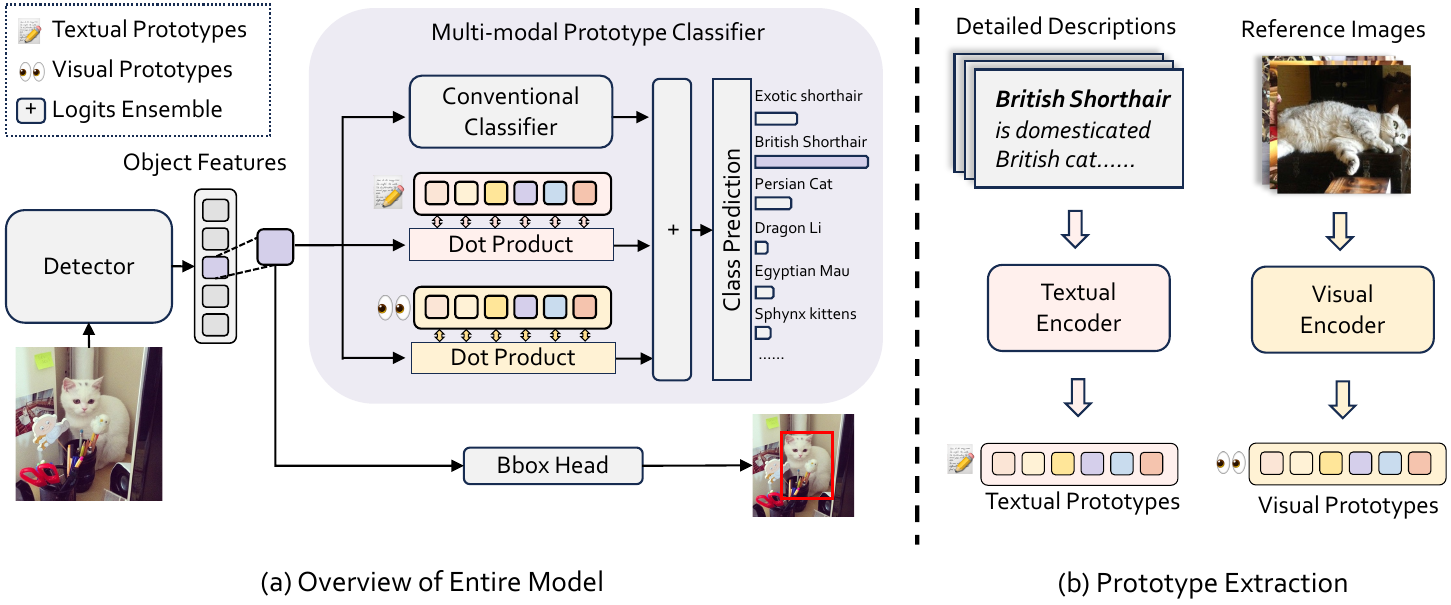} 
\caption{ (a) Overview of Entire Model. Images are processed by a detector, \eg Faster R-CNN~\cite{fastrcnn}, FCOS~\cite{FCOS} or DINO~\cite{dino} to extract object features (RoI features for Faster R-CNN, class features for FCOS and object queries for DINO), then these object features are utilized by the bounding box prediction head to predict bounding boxes and the Multi-modal Prototypes Classifier to generate corresponding category predictions. (b) Prototype Extraction. Textual prototypes are encoded by the LongCLIP-Large~\cite{longclip} text encoder using detailed descriptions from the V3Det Challenge~\cite{v3detcha}, and visual prototypes are encoded by the CLIP-ViT-Large~\cite{clip} visual encoder using images from V3Det~\cite{v3det} training dataset and examples. }
\label{fig:arch}
\end{figure*}
\subsection{Open-Vocabulary Object Detection}
Leveraging the generalization capabilities of Vision-Language Models~\cite{clip, align}, open-vocabulary detectors are able to detect novel classes which are unseen during training. 
ViLD~\cite{vild} effectively distills the knowledge of CLIP into a closed-set detector. MDETR~\cite{mdetr} and GLIP~\cite{glip} convert detection task as a grounding task to expand the training dataset. RegionCLIP~\cite{regionCLIP} uses image-caption data to construct region-wise pseudo-labels as extra training data. Detic~\cite{detic} makes use of image classification data to provide weak supervision. F-VLM~\cite{fvlm} directly builds a two-stage detector upon frozen CLIP and trains the detection heads only, and CORA~\cite{cora} adapts CLIP by region prompting and anchor pre-matching. 

\subsection{Prototype-based Object Recognition}
Prototype is early studied in few-shot and zero-shot learning~\cite{proto}, and is also widely used in other fields~\cite{xing_semi, zhang_prototypical, multi_seg, segic}. In object detection, the vision-language alignment classifiers can be treated as prototypes-based classifier with cosine distance. Besides that, FSOD$^{up}$~\cite{universal} learns universal prototypes as condition to enhance the object classifier. GPA~\cite{GPA} propose a graph-induced prototype alignment framework for cross-domain detection. RPA~\cite{rpn} separately aligns foreground and background RPN features by RPN prototype alignment method. In particular, MM-OVOD~\cite{MMC} employs a method similar to ours by using a multi-modal classifier to enhance open-vocabulary performance. However, unlike MM-OVOD, we do not train additional modules. Instead, we analyze recognition problems from the perspective of vast-vocabulary detection and prototype learning, thereby enhancing the performance in both supervised and open-vocabulary settings.

\section{Method}

\subsection{Preliminaries}\label{sec3.1}
Object detection can be conceptually divided into two parts: object localization and object classification. The former identifies the position of the object within the image by generating a bounding box, while the latter assigns a category label to the localized object.

Traditional detection models~\cite{fastrcnn,FCOS,detr,deformable_detr} usually obtain object features from the entire image according to locations~\cite{fastrcnn,retina,FCOS} or cross-attention mechanism~\cite{detr,deformable_detr,dino}. Then, the object features are fed into a classification branch, \eg MLP, to classify the region into predefined categories, and a location branch to compute the bounding box, separately. Formally, we denote the detector as $D$, the classification branch as $C$, and the localization branch as $L$. Then, these processes can be expressed as:
\begin{equation}
    X = \mathcal{D}(\mathcal{I})
\end{equation}
\begin{equation}
    B = \mathcal{L}(X),\hspace{2mm} S = \mathcal{C}(X)
\end{equation}
where $X$ denotes object features, $B$ and $S$ denotes bounding box predictions and classification logits, separately.

\subsection{Model Overview}

Our detection model primarily comprises a detector, a bounding box (Bbox) prediction head, and a classifier. As illustrated in~\cref{fig:arch} (a), the image is processed by a detector to extract object features. Subsequently, these object features are utilized by the Bbox head to predict the coordinates of the Bbox and the multi-modal classifier to generate the corresponding category predictions. To enable the detection model to handle vast vocabularies in real-world scenarios, we utilize prior knowledge from both the language and visual modalities to initialize two additional alignment classifiers, and ensemble them as an entire multi-modal prototype classifier by averaging their logits.


\subsection{Prototype Extraction}
\subsubsection{Textual Prototypes.}
As shown in~\cref{fig:visual} (a), the textual prototypes generated by adding simple category names to fixed templates possess weak semantics and often fail to maintain a mapping relation with the vast image categories. To address this issue, as illustrated in~\cref{fig:arch} (b), we utilize more detailed descriptive texts for each category from the V3Det Challenge~\cite{v3detcha}, which were obtained by prompting GPT-4V~\cite{2023GPT4VisionSC}. These descriptions are encoded by a text encoder to produce more semantically rich embedding. We denote the descriptions as $\mathcal{T}$ and text encoder as $\mathbf{Enc}_t$. The description embedding $T$can be extracted by:
\begin{equation}
    T = \mathbf{Enc}_t(\mathcal{T}) : \mathbb{R}^{C \times L_t}
\end{equation}
where $C$ denotes the number of categories and $L_t$ represents the latent dimension of description embedding.
As shown in~\cref{fig:visual} (b), detailed descriptions create a cleaner feature space compared to single vocabulary, and we use these description embedding as textual prototypes to guide the classification of the detection model.

\subsubsection{Visual Prototypes.}
As shown in~\cref{fig:arch} (b), we acquire reference images from example images of each category in V3Det, as well as images cropped from the ground-truth in the training set. For each category $c$, 
\begin{equation}
    V^{c} = \mathbf{Enc}_v(\mathcal{I}_r^c) : \mathbb{R}^{n \times L_v}
\end{equation}
where $n$ represents the number of reference images for this category, and $L_v$ represents the latent dimension of image embedding. We then perform weighted fusion based on the image resolution size which can be formally expressed by:
\begin{equation}
    V = \mathcal{W} \cdot V^{c} : \mathbb{R}^{C \times n} \times \mathbb{R}^{n \times L_v} \rightarrow \mathbb{R}^{C \times L_v}
\end{equation}
where $\mathcal{W}$ denotes weighted matrix based on the image resolution size, $C$ denotes the number of categories and $L_v$ represents the latent dimension of image embedding.

We utilize CLIP-ViT-Large as our visual encoder, and the embedding $V$ as visual prototypes to guide the classification of the detection model. As shown in~\cref{fig:visual} (c), the feature space of the visual modality exhibits greater distinction and less ambiguity compared to that of the language modality.

\subsection{Multi-modal Prototype Classifier}

\subsubsection{Conventional Classifier.} 
Let the object features be $X \in \mathbb{R}^{N \times L}$ , where $N$ represents the number of objects (\ie, the number of anchors or queries) and $L$ represents the latent dimension. The classifier can be expressed as follows:
\begin{equation}
    S_{con} = W \cdot X : \mathbb{R}^{C \times L} \times \mathbb{R}^{L \times N} \rightarrow \mathbb{R}^{C \times N}
\end{equation}
Where $S_{con} \in \mathbb{R}^{C \times N}$ denotes the classification logits, $W \in \mathbb{R}^{C \times L}$ denotes the learnable classifier parameters, and $C$ denotes the number of categories. 

We can also consider the conventional closed-set classifier as a specialized alignment method that learns abstract prototypes $W$ and performs classification by calculating the similarity between $W$ and $X$. Given that $W$ lacks any prior knowledge and is initialized randomly, it converges slowly, exhibits weak generalization capability, and poses challenges when expanding to new categories.

\subsubsection{Textual Prototype Classifier.}
Textual prototype classifier employs textual prototypes as prior knowledge to perform alignment and then classifies the object based on cosine similarity. Specifically, let the textual prototype be $T \in \mathbb{R}^{C \times L_t}$, $C$ denotes the number of categories and $L_t$ represents the latent dimension of textual prototypes. The textual prototype classifier can be expressed as follows:

\begin{equation}
    S_{text} = \tau \cdot \frac{T \cdot (P_t \cdot X)}{||T|| \cdot ||P_t \cdot X||} 
\end{equation}
Where $S_{text} \in \mathbb{R}^{C \times N}$ denotes the classification logits by textual prototype classifier, $P_t \in \mathbb{R}^{L_t \times L}$ represents the projection layer that maps the latent dimension of $X$ to the latent dimension of $T$, $\tau$ denotes the temperature of logits scale.

\begin{table*}[!ht]
      \centering
      \tablestyle{15pt}{1.1}
      \begin{tabular}{l | c | c | l  l  l}
      \toprule
      Method & Backbone & Epochs & $AP$ & $AP_{50}$ & $AP_{75}$\\
      \midrule

      Faster R-CNN$^\star$ & ResNet-50 & 24 & 25.4 & 32.9 & 28.1  \\

      Faster R-CNN$^\star$ + \system & ResNet-50 & 24 & \textbf{28.7} (+3.3) & \textbf{37.1} (+4.2) & \textbf{31.7} (+3.6)\\
      
      \hline
      FCOS$^\star$ & ResNet-50 & 24 & 9.4 & 11.7 & 10.1\\
      FCOS$^\star$ + \system & ResNet-50 & 24 & \textbf{15.6} (+6.2) & \textbf{19.1} (+7.4) & \textbf{16.7} (+6.6) \\
      \hline
      Deformable DETR  & ResNet-50 & 50 & 34.4 & 39.9 & 36.4 \\
      DINO & ResNet-50 & 24 & 33.5 & 37.7 & 35.0 \\
      DINO + \system & ResNet-50 & 24 & \textbf{36.4} (+2.9) & \textbf{41.3} (+3.6) & \textbf{38.1} (+3.1)\\
      \midrule
      Faster R-CNN$^\star$ & Swin-Base & 24 & 37.6 & 46.0 & 41.1  \\
      FCOS$^\star$ & Swin-Base & 24 & 21.0 & 24.8 & 22.3 \\
      Deformable DETR  & Swin-Base & 50 & 42.5 & 48.3 & 44.7 \\
      DINO & Swin-Base & 24  & 42.0 & 46.8 & 43.9 \\
      DINO + \system & Swin-Base & 24 & \textbf{44.5} (+2.5)& \textbf{49.9} (+3.1) & \textbf{46.6} (+2.7)\\
      \bottomrule
      \end{tabular}
    \caption{ Main results of supervised object detection on V3Det validation set. All the backbones are pre-trained on ImageNet-1K~\cite{imagenet}. $^\star$ indicates utilizing Norm Linear Layer in the classification head~\cite{normlinearlayer}. }
    \label{tab:main}
\end{table*}

\subsubsection{Visual Prototype Classifier.}
Similar to the textual prototype classifier, the visual prototype classifier also employs alignment method to enhance classification performance by utilizing visual prototypes. Specifically, let the visual prototype be $V \in \mathbb{R}^{C \times L_v}$, $C$ denotes the number of categories and $L_v$ represents the latent dimension of visual prototypes. The visual prototype classifier can be expressed as follows:

\begin{equation}
    S_{vis} = \tau \cdot \frac{V \cdot (P_v \cdot X)}{||T|| \cdot ||P_t \cdot X||} 
\end{equation}
Where $S_{vis} \in \mathbb{R}^{C \times N}$ denotes the classification logits by visual prototype classifier, $P_v \in \mathbb{R}^{L_v \times L}$ represents the projection layer that maps the latent dimension of $X$ to the latent dimension of $V$, $\tau$ also denotes the temperature of logits scale.

\begin{table*}[!ht]
      \centering
      \tablestyle{10pt}{1.0}
      \begin{tabular}{c | c | c | c | c| c | c | c}
      \toprule
      Method & Backbone & Params &$\mathcal{D}^{backbone}$ & $\mathcal{D}^{detector}$ & $AP$ & $AP_{50}$ & $AP_{75}$\\
      \midrule
      Cascade Mask R-CNN & EVA-ViT-G & 1.1B & Merged-30M & Object365 & 49.4 & 54.8 & 51.4 \\
      DINO + \system & Swin-Base & 213.4M & ImageNet-22K & $\times$ & \textbf{50.3} & \textbf{56.1} & \textbf{52.6} \\
      \bottomrule
      \end{tabular}
    \caption{ Final results of supervised object detection on V3Det validation set. Params denotes the size of entire models. $\mathcal{D}^{backbone}$ indicates the pre-training dataset of backbones and $\mathcal{D}^{detector}$ indicates the pre-training dataset of detectors.  Merged-30M~\cite{eva} is a mixed dataset including ImageNet-22K, CC12M, CC3M, Object365, COCO and ADE20K.}
    \label{tab:final}
\end{table*}
Formally, the prototype classifier provides semantically rich initialization matrices $T$ and $V$ to replace the role of $W$ in the conventional classifier. As a result, it achieves better classification performance and faster convergence. Finally, we simply ensemble all classifiers for supervised setting, and remove closed-set classifier for open-vocabulary setting:
\begin{equation}
    S = 
    \begin{cases}
    \begin{array}{cc}
    \frac{S_{con} + S_{text} + S_{vis}}{3} & \text{if supervised} \\[2pt]
    \frac{S_{text} + S_{vis}}{2} & \text{else}
    \end{array}
    \end{cases}
\end{equation}

\section{Experiments}
In this section, we evaluate our method with various baselines on V3Det validation set in both supervised and open-vocabulary manners. We conduct additional experiments on LVIS in an open-vocabulary setting. Finally, we present ablation studies on different components of our classifier.

\subsection{Supervised Object Detection}
\subsubsection{Datasets.} We conduct experiments on V3Det, a vast-vocabulary object detection dataset containing 13204 categories with $\sim$200K images. We train our models with all 13204 categories in V3Det train set with $\sim$180K images, and evaluate them on the validation set with $\sim$30K images.

\subsubsection{Baselines.}
We consider the following baselines: (1) Faster R-CNN~\cite{fastrcnn}: a two-stage object detector. (2) FCOS~\cite{FCOS}: a one-stage object detector. (3) DINO~\cite{dino}: a DETR-based model. These baselines are to validate our proposed \system. Moreover, we compare our best model with ImageNet-22k pre-trained Swin-Base~\cite{swin} as the backbone 
to the previous SoTA detector with EVA-Huge~\cite{eva}.

\subsubsection{Implementation Details.}
We use 8 $\times$ V100 for all experiments. For Faster R-CNN and FCOS, we follow the setups from V3Det without any modification except halving the batch size of FCOS. For DINO, we set the initial learning rate as 0.0002 with 24 epochs and multiplied 0.1 at the 20th epoch. We obtain descriptions from V3Det Challenge~\cite{v3detcha} and collect reference images from example images and training set in V3Det no more than 100 per category. The object features used to recognize objects are RoI features for Faster R-CNN, dense grid features for FCOS, and object queries for DINO. Following V3Det~\cite{v3det}, we adopt repeat factor sampler~\cite{lvis} and report average precision under different IoU noted as $AP$, $AP_{50}$, $AP_{75}$.

\begin{table*}[!ht]
      \centering
      \tablestyle{11pt}{1.0}
      \begin{tabular}{c | c | c | c | c | c | c}
      \toprule
      Method & Backbone & Extra Training Data &Epochs & $AP_{base}$ & $AP_{novel}$ & $AP_{final}$\\
      \midrule
      
      DINO$^\ast$ & ResNet-50 & $\times$ & 24 &  19.0 & 1.9 &  10.5\\
      RegionCLIP  & ResNet-50 & $\times$ & 48 &  22.1 & 3.1 & 12.6 \\
      CenterNet2$^\ast$ & ResNet-50 & $\times$ & 48 & 28.6 & 3.0 & 15.8 \\
      DST-Det & ResNet-50 & $\times$ & 24 &  \MissedData & 7.2 & \MissedData \\
      DINO + \system & ResNet-50 & $\times$ & 24 &  \textbf{31.4} & \textbf{9.5} & \textbf{20.5}  \\
      \midrule
      Detic & ResNet-50 & ImageNet-22K & 48 & 30.2 & 6.7 & 18.5 \\
      DINO + \system & ResNet-50 & ImageNet-22K & 24 & \textbf{32.8} & \textbf{11.0} & \textbf{21.9} \\
      \bottomrule
      \end{tabular}
    \caption{ Results of open-vocabulary object detection on V3Det validation set. $^{\ast}$ indicates replacing closed-set classifier with class name alignment classifier; - indicates that do not have reported number. $AP_{final}$ is the mean of $AP_{base}$ and $AP_{novel}$.}
    \label{tab:final_ovd}
\end{table*}

\begin{table*}[!ht]
      \centering
      \tablestyle{11pt}{1.0}
      \begin{tabular}{c | c | c | c | c | c | c}
      \toprule
      Method & Backbone & Backbone Size &$\mathcal{D}^{backbone}$ & V-L Pretraining & $AP_{r}$ & $AP$\\
      \midrule
      RegionCLIP  & ResNet-50x4 & 87M & CLIP-400M & \checkmark &  22.0$^{\ast}$ & 32.3 \\
      CORA  & ResNet-50x4 & 87M & CLIP-400M & \checkmark &  22.2 & - \\
      CLIPSelf  & ViT-B/16 & 86M & Merged-2B & \checkmark &  25.3$^{\ast}$ & - \\
      RO-ViT  & ViT-B/16 & 86M & ALIGN & \checkmark &  28.4 & 31.9 \\
      CFM-ViT  & ViT-B/16 & 86M & ALIGN & \checkmark &  29.6 & 33.8 \\
      DINO + \system & Swin-Base & 88M & ImageNet-22K & $\times$ & \textbf{31.5} & \textbf{44.2} \\
      \bottomrule
      \end{tabular}
    \caption{ Results of open-vocabulary object detection on OV-LVIS. $\mathcal{D}^{backbone}$ indicates pre-training data of backbone; V-L Pretraining indicates whether using contrastive image-text pre-training; $^{\ast}$ indicates reporting mask AP.}
    \label{tab:lvis}
\end{table*}

\subsubsection{Results.}

\cref{tab:main} shows the result on V3Det val set for supervised setting, and all the backbones pre-trained on ImageNet-1K~\cite{imagenet} following V3Det. As shown in this table, when using a ResNet-50~\cite{resnet} as the backbone, different frameworks consistently benefit from our proposed \system (\eg, +3.3 AP on Faster R-CNN, +6.2 AP on FCOS and +2.9 AP on DINO). These results demonstrate that \system is a general method applicable to popular detection frameworks. When using a Swin-Base~\cite{swin} as the backbone, our method largely outperforms the baselines (\eg, +2.5 AP on DINO), suggesting that \system is also an efficient method and suitable for larger models. 

Moreover, we compare our biggest model, DINO + \system with Swin-Base pre-trained on ImageNet-22K~\cite{imagenet}, to a strong Cascade Mask R-CNN detector, which uses EVA-ViT-G~\cite{eva, vit} as backbone network pre-trained on Merged-30M, which includes ImageNet-22K, CC12M~\cite{cc12m}, CC3M~\cite{cc3m}, Object365~\cite{objects365}, COCO~\cite{coco}, and ADE20K~\cite{ade}. Then the detector is pre-trained again on Object365. As shown in~\cref{tab:final}, despite the model being five times smaller and under a much weaker pre-training strategy, \system outperforms this detector by 0.9 AP, demonstrating its efficiency and remarkable performance.

\subsection{Open-vocabulary Object Detection}

\subsubsection{Datasets.}
We further experiments on V3Det under an open-vocabulary setting, where the classes are split into 6709 base classes and 6495 novel classes. We train our models on base classes with $\sim$130K images, and evaluate them on the validation set including both base classes and novel classes. We also utilize extra data from ImageNet-22K which overlaps the vocabularies of V3Det.

\subsubsection{Baselines.}
We choose four different models:
(1) RegionCLIP~\cite{regionCLIP} trained on base classes for 24 epochs to initialize the offline RPN and 24 epochs again to align regions and texts. (2) CenterNet2~\cite{center} trained on V3Det base classes. (3) DST-Det~\cite{dst} trained on V3Det base classes. (4) Detic~\cite{detic} trained on both V3Det base classes and ImageNet-22K overlapped classes. The backbones of four models are pre-trained on ImageNet-22k. We also report DINO with class name alignment classifier as the baseline.

\subsubsection{Implementation detail.}
We utilize DINO with \system as our open-vocabulary model. The backbone network is also pre-trained on ImageNet-22k. For mixed dataset training, we follow the training recipe in RichSem~\cite{richsem} to sample images from V3Det and ImageNet-22K in a 1:1 ratio; for data from ImageNet-22k, we treat a whole-image box as ground-truth bounding box and group four images together to build a mosaic following~\cite{richsem}. Again, we set the initial learning rate as 0.0002 with 24 epochs and multiply 0.1 at the 20-th epoch. The textual prototypes and base class visual prototypes are the same as ones in the supervised setting. In contrast, for novel classes, we obtain reference images from example images in V3Det Challenge~\cite{v3detcha} 1 image per category and ImageNet-22K 100 images per category if overlapped, to ensure that no novel images from the V3Det training set are seen.

\begin{table*}[!ht]
    \centering
    \tabcolsep=0.15cm
      \tablestyle{12pt}{1.0}
    
    \begin{tabular}{*{4}{c}|*{3}{c}}
        \toprule
        \makecell{Class-name \\ Alignment} &
        \makecell{Description \\ Alignment} &\makecell{Visual \\ Alignment} &  \makecell{Conventional \\ Classifier} & $AP$ & $AP_{50}$ & $AP_{75}$\\
        \midrule
        \checkmark & & & & 10.8 & 12.9 & 11.5\\
         & \checkmark & & & 16.0 & 18.7 & 16.8 \\
         &  & \checkmark & & 28.1 & 32.2 & 29.5 \\

         &  & & \checkmark & 24.5 & 28.2 & 25.7 \\
        \hline
        &  \checkmark & \checkmark &  & 29.4 & 33.6 & 30.9\\
        &  \checkmark & \checkmark & \checkmark & 34.0 & 38.8 & 35.6\\
        \bottomrule
        
    \end{tabular}   
    \caption{Ablation on different classifiers. If utilizing multiple classifiers, we ensemble them by averaging the prediction logits.}
    \label{tab:ablation}
\end{table*}

\subsubsection{Results.}
In~\cref{tab:final_ovd}, \system largely outperforms the open-vocabulary models whether or not to use extra training data. Compared to the strong baseline models, \ie CenterNet2/Detic~\cite{detic, v3det}, \system obtain +4.3 $AP_{novel}$ and +3.4 $AP_{final}$ when using extra data, and +6.5 $AP_{novel}$ and +4.7 $AP_{final}$ without extra data, demonstrating that our model gains strong generalization capability.

\subsubsection{Additional Dataset.}
We evaluate \system on LVIS~\cite{lvis} which has 1203 categories. We simply extract visual prototypes by less than 100 images cropped from the LVIS training-set, and directly treat the class-name embedding as the textual prototypes. As shown in \cref{tab:lvis}, unlike other detectors~\cite{regionCLIP, cora, clipself, rovit, cfmvit} utilizing vision-language pretraining into model design, our offline prototype-based method effectively achieves a new state-of-the-art performance under the base-size comparison.

\subsection{Ablation Study}
The evaluations are performed on V3Det dataset under supervised training of 12 epochs with DINO as detector and ResNet-50 pre-trained on ImageNet-1K as backbone. 

\subsubsection{Effectiveness of Different Classifier.}

We evaluate the effects of each component within the multi-modal prototype classifier. As shown in \cref{tab:ablation}, compared to coarse class names, detailed descriptions enhance the differentiation between categories, and visual prototypes have greater distinction and reduced ambiguity. Integrating both prototypes, the classifier can better align with the detected objects.

\begin{figure}[!t]
\centering
\includegraphics[width=0.47\textwidth]{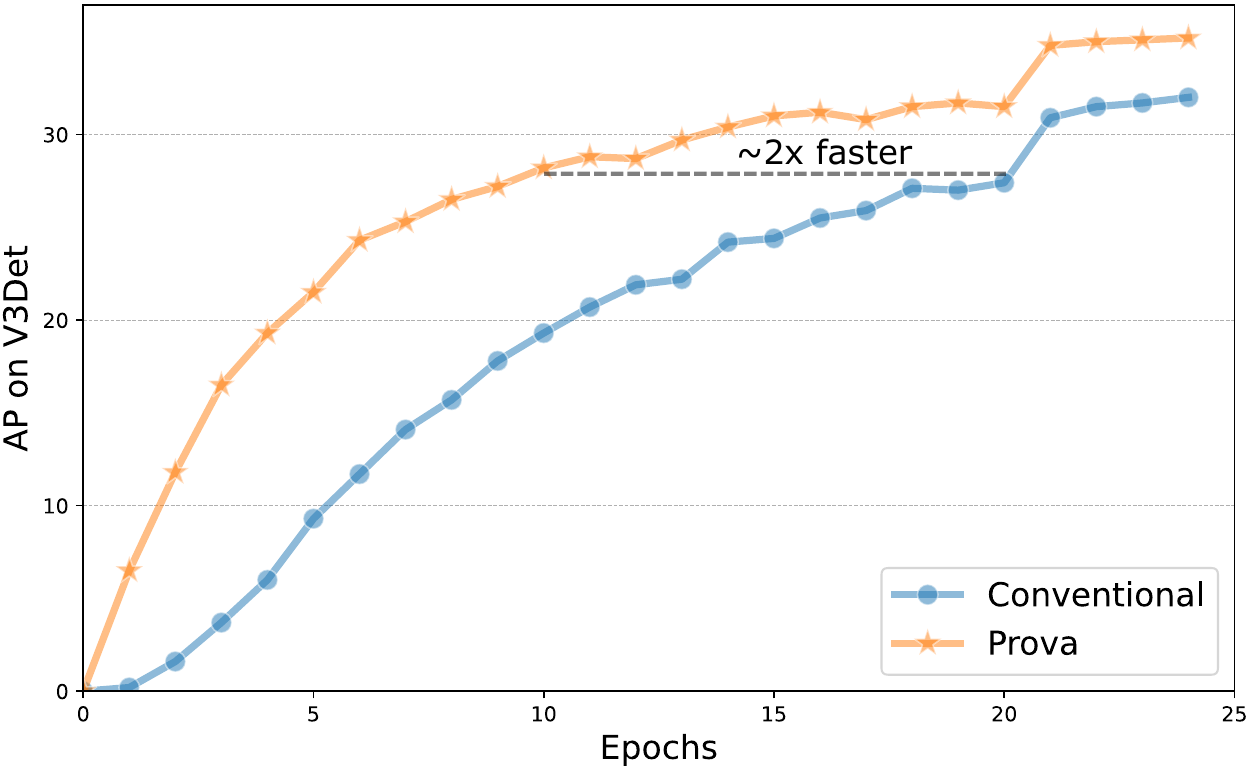} 
\caption{ Convergence curves of DINO with conventional classifier and \system for 24 epochs on V3Det val set.}
\label{fig:convergence}
\end{figure}

Additionally, we plot the AP curves of \system and conventional closed-set classifier trained for 24 epochs in a supervised setting. As show in~\cref{fig:convergence}, \system has faster convergence with multi-model prototypes compared to the conventional classifier, suggesting the efficiency of our method.

\subsubsection{Comparisons with Different Encoders of prototypes.}

As shown in~\cref{tab:ablation2}, we explore the visual prototype encoder with CLIP-ViT-Large, SigLIP-Large~\cite{siglip} and DINOv2-Large~\cite{dinov2}, finding that better results are achieved when utilizing CLIP, so we adopt it as our default visual encoder. In Addition, we also evaluate the textual prototype encoder with LongCLIP-Large, CLIP-ViT-Large and T5~\cite{t5}. As shown in~\cref{tab:ablation3}, LongCLIP achieves better performance owing to its compatibility with long text input and training objective.

\subsubsection{Effectiveness of the Number of Reference Images.}
We study the impact of visual prototype quality on the recognition results by using varying numbers of reference images. As shown in~\cref{tab:ablation4}, the more reference images lead the better the result, suggesting that a greater number of reference images can result in more balanced visual prototypes.

\begin{table}[!ht]
      \centering
      \tablestyle{6pt}{1.0}
      \begin{tabular}{c | c | c | c}
      \toprule
      Model & $AP$ & $AP_{50}$ & $AP_{75}$\\
      \midrule
      CLIP-ViT-Large & 28.1  &  32.2 & 29.5 \\
      SigLIP-Large & 27.5 & 31.5 & 28.8 \\
      
      DINOv2-Large & 25.5 & 29.4 & 26.8 \\
      \bottomrule
      \end{tabular}
    \caption{Ablation on visual prototype encoders.}
    \label{tab:ablation2}
\end{table}

\begin{table}[!ht]
      \centering
      \tablestyle{6pt}{1.0}
      \begin{tabular}{c | c | c | c}
      \toprule
      
      Model & $AP$ & $AP_{50}$ & $AP_{75}$\\
      \midrule

      LongCLIP-Large & 16.0  &  18.7 & 16.8 \\
      CLIP-ViT-Large$^\dagger$ & 11.9  &  14.0 & 12.6 \\
      T5-3B   & 8.4  &  10.0 & 8.8 \\
      \bottomrule
      \end{tabular}
    \caption{Ablation on textual prototype encoders. $\dagger$: Due to the maximum text token length, we truncate any part of a description that exceeds the limit.}
    \label{tab:ablation3}
\end{table}

\begin{table}[!ht]
      \centering
      \tablestyle{6pt}{1.0}
      \begin{tabular}{c | c | c | c}
      \toprule
      
      \makecell{Number of \\ reference images} & AP & AP$_{50}$ & AP$_{75}$\\
      \midrule
      1 & 21.3  &  24.4 & 22.2 \\
      
      10 & 26.6 & 30.5 & 27.9 \\
      100 & 28.1  &  32.2 & 29.5 \\
      \bottomrule
      \end{tabular}
    \caption{ Ablation on the number of reference images. }
    \label{tab:ablation4}
\end{table}

\section{Conclusion}
We proposed \textbf{\system}, a simple and effective multi-modal \textbf{pro}totype classifier for \textbf{va}st-vocabulary object detection, to leverage multi-modal prototypes extracted by strong foundation models, alleviating the decline of object recognition performance when training categories scale up. Through extensive experiments, we demonstrated that our approach achieved state-of-the-art performance effectively. 

\section{Acknowledgements} This project was supported by NSFC under Grant No.   62102092.


\clearpage
\newpage
\onecolumn
\appendix

\section{Technical Appendix}
\subsection{Pre-processing of Visual Prototype Extraction}
For the supervised setting, we acquire reference images from example images in V3Det~\cite{v3det}, as well as images cropped from the ground-truth in the training set no more than 100 images per class, and sort the cropped images in descending order by resolution size. For each category $c$, we obtain image embedding $V^{c} \in \mathbb{R}^{n \times L_v}$ through CLIP-ViT-Large~\cite{clip} visual encoder, where $n$ represents the number of reference images for this category, and $L_v$ represents the latent dimension of image embedding. We denote $V^c_i$ is the $i-th$ image embedding in $V^c$, where $i \in [1, n]$, and $V^c_1$ is the embedding of exemplar in category $c$. If $n$ is more than 1, we then perform weighted fusion which can be formally expressed by:
\begin{equation}
    V^c_{fused} = \sigma(n) \cdot V^c_1 + \sum_{i=2}^n \frac{1 - \sigma(n)}{n - 1} \cdot V^c_i
\end{equation}
where the mapping table from n to $\sigma$ is shown in Table \ref{tab:app1}. Finally, we stack all the $V^c_{fused}$ for each category $c$ to obtain visual prototypes $V \in \mathbb{R}^{C \times L_v}$, where $C$ denotes the number of categories.

\begin{table}[!ht]
      \centering
      \tablestyle{20pt}{1.4}
      \begin{tabular}{c | c}
      \toprule
      n & $\sigma(n)$ \\
      \midrule
      1 & 1.0 \\
      2 & 0.6 \\
      3 & 0.5 \\
      4 & 0.4 \\
      5 $\sim$ 7 & 0.3 \\
      8 $\sim$ 10 & 0.2 \\
      11 $\sim$ 20 & 0.15 \\
      21 $\sim$ 50 & 0.12 \\
      51 $\sim$ 100 & 0.10 \\
      \bottomrule
      \end{tabular}
    \caption{Mapping table from $n$ to $\sigma$, where $n$ represents the number of reference images for one category, and $\sigma$ denotes the fusion weight.}
    \label{tab:app1}
\end{table}

For the open-vocabulary setting, we obtain images in ImageNet-22K~\cite{imagenet} for novel classes, ensuring that no more than 100 images are used per category, and have no modification for the fusion recipe mentioned above.

\subsection{Detailed Evaluation Results on OV-LVIS}
We additionally evaluate our method on OV-LVIS which has 1203 categories. We simply extract visual prototypes by less than 100 images cropped from the LVIS training-set, and directly treat the class-name embedding as the textual prototypes.
\begin{table*}[hb!]
      \centering
      \tablestyle{10pt}{1.2}
      \begin{tabular}{c | c | c | c}
      \toprule
      Method & $Backbone $ & ImageNet-22K & $AP_{r}$\\
      \midrule
      DINO w/ Prova (only textual prototype) & Swin-Base  &  $\times$ & 13.2 \\
      DINO w/ Prova (textual + visual prototype) & Swin-Base & $\times$ & 15.4 \\
      DINO w/ Prova (textual + visual prototype) & Swin-Base & \checkmark & 31.5 \\
      \bottomrule
      \end{tabular}
    \caption{Evaluation on OV-LVIS. ImageNet-22K indicates using ImageNet-22K as extra training data.}
    \label{tab:ablation5}
\end{table*}

As shown in \cref{tab:ablation5}, although there exists a big gap of category numbers between LVIS (1203) and V3Det (13204), our method still shows consistent improvement on the rare class on OV-LVIS validation set, indicating the generalizability of our proposed Prova. In addition, this result can be futher improve by methods like treating detailed description embedding as textual prototypes, picking more high-quality reference images, etc.
\end{document}